\documentclass[10pt,twocolumn,letterpaper]{article}

\usepackage{cvpr}      

\definecolor{cvprblue}{rgb}{0.21,0.49,0.74}
\usepackage[pagebackref,breaklinks,colorlinks,allcolors=cvprblue]{hyperref}

\def\paperID{*****} 
\def\confName{CVPR}
\def\confYear{2025}

\title{Enhancing Sa2VA for Referent Video Object Segmentation: 2nd Solution for 7th LSVOS RVOS Track}

\newcommand*{\email}[1]{\tt\small{#1}}

\author{
    Ran Hong\textsuperscript{1,2*} \quad 
    Feng Lu\textsuperscript{3*} \quad 
    Leilei Cao\textsuperscript{1} \quad
    An Yan\textsuperscript{1} \quad
    Youhai Jiang\textsuperscript{1} \quad
    Fengjie Zhu\textsuperscript{1} \\
    \textsuperscript{1}TEX AI, Transsion Holdings \quad
    \textsuperscript{2}Nanchang University \quad
    \textsuperscript{3}ShanghaiTech University \\
    \textsuperscript{*}: Equal technical contribution to the work. \\
    \email{ranhong@email.ncu.edu.cn, lufeng2023@shanghaitech.edu.cn} \\ \email{leilei.cao@transsion.com, an.yan@transsion.com}
}

\begin{document}
\maketitle
\begin{abstract}
Referential Video Object Segmentation (RVOS) aims to segment all objects in a video that match a given natural language description, bridging the gap between vision and language understanding. Recent work, such as Sa2VA, combines Large Language Models (LLMs) with SAM~2, leveraging the strong video reasoning capability of LLMs to guide video segmentation. In this work, we present a \textbf{training-free} framework that substantially improves Sa2VA's performance on the RVOS task. Our method introduces two key components: (1) a \textbf{Video-Language Checker} that explicitly verifies whether the subject and action described in the query actually appear in the video, thereby reducing false positives; and (2) a \textbf{Key-Frame Sampler} that adaptively selects informative frames to better capture both early object appearances and long-range temporal context. Without any additional training, our approach achieves a $\mathcal{J}\&\mathcal{F}$ score of 64.14\% on the MeViS test set, ranking \textbf{2nd place} in the RVOS track of the 7th LSVOS Challenge at ICCV 2025.
\end{abstract}
    
\section{Introduction}

Referring Video Object Segmentation (RVOS) aims to segment the target object in all frames of a video based on a natural language description. Given a video and a textual query such as \textit{``the man in the red shirt''}, the model must produce a sequence of temporally consistent masks for the referred object. Compared to image-level referring segmentation, RVOS is considerably more challenging due to three factors: (1) the target's appearance may change drastically across frames because of motion, occlusion, and lighting variations; (2) temporal consistency must be preserved to avoid mask flickering and object drifting; and (3) robust vision-language alignment must be maintained throughout the entire video.

The importance of RVOS has been underscored by its inclusion as one of the official tracks in the 7th Large-Scale Video Object Segmentation (LSVOS) Challenge. The challenge features three tracks: (1) \textbf{Complex Video Object Segmentation (MOSEv2)} \cite{ding2025mosev2challengingdatasetvideo}, which targets challenging scenarios involving frequent object disappearance and reappearance, severe occlusions, small targets, and adverse conditions such as low light, multi-shot sequences, camouflage, and knowledge-dependent reasoning; (2) \textbf{Classic VOS} \cite{ding2023mosenewdatasetvideo}, which segments a specific object instance throughout the entire video given only its first-frame mask, using a mixture of MOSEv1 and LVOS datasets; and (3) \textbf{Referring VOS (RVOS)} \cite{ding2023mevislargescalebenchmarkvideo,ding2025mevis}, which focuses on segmenting objects according to natural language expressions. Together, these tracks provide a comprehensive benchmark for advancing both instance-specific and language-driven video object segmentation methods.

Early RVOS approaches typically followed a two-stage paradigm, first generating candidate regions and then matching them to the textual query. However, such methods incurred high computational cost and were prone to error propagation. Inspired by DETR \cite{detr}, recent work reformulates RVOS as a query-based set prediction problem. MTTR \cite{mttr} introduced a transformer-based architecture that jointly encodes visual and linguistic features and directly predicts pixel-level masks end-to-end. Building on this, ReferFormer \cite{referformer} explicitly models temporal relations through temporal attention and multimodal cross-attention, greatly improving temporal mask consistency. More recently, Refer-DINO \cite{referdino} \cite{cao2022secondplacesolution4th} leverages stronger feature backbones and query refinement strategies to further enhance multimodal grounding and segmentation accuracy.

The rise of foundation models has unlocked new opportunities for RVOS. Large multimodal models (LMMs)  such as InternVL \cite{intervl} and LLaVA \cite{LlaVa} exhibit strong open-vocabulary reasoning capabilities, providing rich semantic priors to disambiguate complex referring expressions. Likewise, large-scale segmentation foundation models such as SAM and its successor SAM~2 \cite{sam2} demonstrate impressive zero-shot segmentation capabilities. When combined with language grounding, these models can produce high-quality masks with minimal fine-tuning, enabling scalable solutions for few-shot and open-world RVOS scenarios.

Despite these advances, RVOS remains difficult in scenarios involving severe occlusion, small targets, or complex natural language queries. Most existing approaches do not explicitly verify whether the described subject and action actually occur in the video, leading to false-positive segmentations when the query is mismatched. Furthermore, uniformly sampling a fixed number of key frames often fails to capture the relevant subject or action, particularly in long videos where the target may appear late.

To overcome these challenges, we propose a novel RVOS framework that integrates large multimodal models with segmentation foundation models. Our approach comprises three components: (1) a \textbf{Video-Language Checker (VLC)} that employs a powerful LMM to determine whether the subject and action described in the text appear in the video, thereby reducing unnecessary computation and false positives; (2) a \textbf{Key-Frame Sampler (KFS)} that combines head-frame sampling with uniform sampling to capture both early appearances and global temporal context, while keeping the number of frames manageable to preserve the representation capacity of the SEG token; and (3) an enhanced Sa2VA \cite{Sa2VA} module that leverages the sampled frames and referring expression to learn a discriminative SEG token, which is then used as a prompt for SAM~2 to segment the key frames and propagate the masks across the entire video.

\noindent\textbf{Our contributions} are summarized as follows:
\begin{itemize}
    \item We introduce a novel RVOS framework that explicitly verifies semantic consistency between the referring expression and the video, effectively reducing false positives when the query does not match the video content.
    \item We propose an adaptive KFS that balances capturing early object appearances with maintaining global temporal coverage, thereby improving SEG token representation and segmentation accuracy.
    \item We enhance the Sa2VA module by learning a more discriminative SEG token from sampled frames and using it as a prompt for SAM~2 to segment key frames and propagate masks across the entire video.
    \item Extensive experiments on the MeViS benchmark demonstrate that our approach achieves state-of-the-art performance, particularly in scenarios where the target appears late or is described by complex action-related expressions.
\end{itemize}

\section{7th LSVOS Challenge Overview}

The 7th Large-Scale Video Object Segmentation (LSVOS) Challenge consists of three tracks, each targeting different aspects of video object segmentation and providing a comprehensive benchmark for advancing VOS research.

\subsection{Complex Video Object Segmentation (MOSEv2)}  
MOSEv2 focuses on video object segmentation in complex real-world scenarios. Videos in this track feature frequent object disappearance and reappearance, severe occlusions, small targets, and additional challenges such as adverse weather (e.g., rain, snow, fog), low-light conditions (e.g., nighttime, underwater), multi-shot sequences, camouflaged objects, non-physical targets (e.g., shadows and reflections), and scenarios requiring external knowledge. The dataset contains 5,024 videos with over 701,976 high-quality masks for 10,074 objects across 200 categories. Benchmarks on representative VOS methods show substantial performance drops compared to simpler datasets, demonstrating the difficulty of real-world scenarios.

\subsection{Classic Video Object Segmentation (Classic VOS)}  
Classic VOS represents the standard video object segmentation setting, where a single object instance must be segmented throughout the video, given only its mask in the first frame. The dataset is a combination of MOSEv1 and LVOS, providing a mixture of familiar and slightly more challenging scenarios. This track evaluates methods under more realistic conditions compared to existing benchmarks like DAVIS and YouTube-VOS, which mainly contain salient, isolated objects.

\subsection{Referring Video Object Segmentation (RVOS)}  
RVOS aims to segment objects in videos guided by natural language expressions, with a particular focus on motion cues. Existing referring video object datasets typically emphasize static attributes, allowing targets to be identified in a single frame. To investigate motion-based language grounding, the MeViS dataset was introduced, containing numerous motion expressions in complex environments. Benchmarks of current RVOS methods on MeViS indicate that existing approaches struggle to leverage motion expressions effectively, highlighting the need for methods that can understand both the semantic content of the text and the temporal dynamics of the video.  

These challenges motivate the design of our framework, which integrates a Video-Language Checker, an adaptive Key-Frame Sampler, and an enhanced Sa2VA module to address semantic verification, key-frame selection, and temporally consistent segmentation.

\section{Method}
\label{sec:method}

Consider a video sequence consisting of $T$ frames $\mathcal{V} = \{\mathbf{I}_t\}_{t=1}^{T}$, where each frame $\mathbf{I}_t \in \mathbb{R}^{3 \times H \times W}$ represents an RGB image of height $H$ and width $W$. Given a referring expression $\mathcal{T} = \{w_i\}_{i=1}^{L}$, where $w_i$ denotes the $i$-th token, the goal of RVOS is to predict a sequence of binary masks $\mathcal{M} = \{\mathbf{M}_t\}_{t=1}^{T}$ with $\mathbf{M}_t \in \{0,1\}^{H \times W}$, such that $\mathbf{M}_t$ identifies the spatial region of the object referred to by $\mathcal{T}$ in frame $\mathbf{I}_t$. Formally, RVOS can be defined as learning a mapping $f: (\mathcal{V}, \mathcal{T}) \mapsto \mathcal{M}$.

\begin{figure*}[t] 
    \centering
    \includegraphics[width=\textwidth]{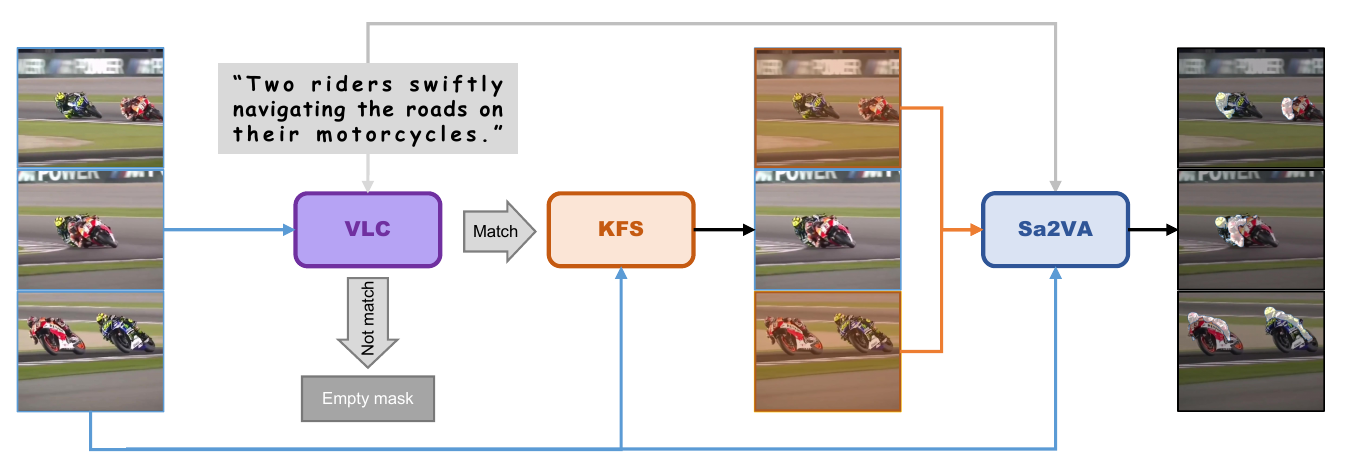}
    \caption{Overall framework of our method.}
    \label{fig:overall}
\end{figure*}

As shown in the Figure \ref{fig:overall}, Our proposed approach for RVOS is composed of three sequential modules.

\textbf{VLC (Section 2.1)} evaluates the correspondence between the input video and the referring expression, ensuring that subsequent segmentation is performed only when the text matches the video content.

\textbf{KFS (Section 2.2)} selects a set of informative key frames from the video to reduce temporal redundancy and provide focused input for the segmentation module.

\textbf{Sa2VA (Section 2.3)} takes the sampled key frames, the preceding video frames, and the referring expression as input, and sequentially predicts the binary masks of the referred object in each frame. This modular design enables our framework to first verify textual-visual relevance, focus on informative frames, and finally generate temporally consistent segmentation masks aligned with the referring expression.

\subsection{Video Semantic Matching}

The RVOS task can be understood as segmenting the object in a video that matches a given textual description. This matching involves both subject-level correspondence and action-level correspondence. If the video does not contain the subject described in the text, or if the subject's action does not correspond to the described action, a zero mask should be output.

To address this, we employ a large pre-trained model from QwenVL \cite{QwenVL} to perform video-text correspondence verification. Specifically, in the VLC module, the video sequence and the corresponding referring expression are input to the QwenVL along with the prompt:

"Please check whether the video matches the input text, i.e., whether the subject described in the text exists in the video and whether the subject's action corresponds to the action described in the text. Output yes/no."

If the model outputs no, all segmentation masks are set to zero. Otherwise, the video proceeds to subsequent modules for key-frame sampling and object segmentation.

\subsection{Key Frame Sampler}

Compared to conventional video object segmentation (VOS) tasks, the RVOS task poses additional challenges due to the need for semantic understanding and precise alignment between the textual description and video content.

The original Sa2VA model is a multi-task framework not specifically designed for RVOS. In its default configuration, Sa2VA selects the first five frames as key frames. However, for datasets such as MeVis, the object corresponding to the textual description may not appear in the first few frames, and action understanding often requires observing more frames. In Sa2VA, the key frames and the referring expression are input into a large pre-trained model to learn a SEG token, which is then used to segment the object in the key frames. Since the output is always a single SEG token regardless of the number of key frames, selecting too many frames can reduce the expressive capacity of the SEG token. Therefore, a reasonable key-frame selection strategy is critical to improving segmentation accuracy.

A straightforward approach is uniform sampling across the video, effectively compressing the original video. However, uniform sampling suffers from several limitations: the optimal number of sampled frames is difficult to determine, and for long videos, consecutive sampled frames may be too far apart, resulting in a loss of action information. Observing that most objects corresponding to the text appear in the early part of the video, we adopt a hybrid strategy combining head continuous sampling with uniform sampling. This approach controls the number of key frames to maintain the SEG token’s expressive capacity while also capturing action dynamics in longer videos, leading to more accurate segmentation of the referred objects.

\subsection{Sa2VA for segmentation}

After sampling key frames and obtaining the SEG token from the LLM in Sa2VA, the SEG token is used as a prompt for the SAM2 Decoder to segment the object in the key frames. Subsequently, the masks obtained from the key frames are propagated as prompts through SAM2 across the entire video, enabling the segmentation of all frames corresponding to the textual description. This two-stage process—first segmenting the key frames and then propagating the masks—ensures that the model captures both the object appearance and its temporal dynamics, yielding accurate segmentation for all frames in the video.

The entire procedure of our framework can be summarized as follows: for each frame $t$, the segmentation mask $\mathbf{M}_t$ is given by

\[
\mathbf{M}_t =
\begin{cases}
\mathbf{0}, & \text{if } f_\text{VLC}(\mathcal{V}, \mathcal{T}) = 0, \\
f_\text{Sa2VA}(\mathbf{I}_1, \dots, \mathbf{I}_t, \{\mathbf{I}_{t_k}\}_{k=1}^{N}, \mathcal{T}) , & \text{if } f_\text{VLC}(\mathcal{V}, \mathcal{T}) = 1
\end{cases}
\]

where $f_\text{VLC}$ denotes the VLC module that determines whether the video $\mathcal{V}$ contains the subject and action described in the referring expression $\mathcal{T}$. The set $\{\mathbf{I}_{t_k}\}_{k=1}^{N}$ represents the $N$ key frames sampled from the video. The function $f_\text{Sa2VA}$ represents the Sa2VA model, which first segments the key frames using the SEG token as a prompt for the SAM2 decoder and then propagates the masks across the entire video. In this way, each mask $\mathbf{M}_t$ depends only on the first $t$ frames, the sampled key frames, and the referring expression.


\section{Experiment}

\subsection{Dataset and Metrics}
\textbf{Dateset}. MeVis is a large RVOS dataset containing 2K videos along with corresponding target motion description texts. In this competition, 1,662 videos are used for training, 190 videos for validation, and 50 videos for testing.
\textbf{Metrics}. We employ $\mathcal{J}$ (average IoU) to evaluate region similarity, and $\mathcal{F}$ (mean boundary similarity) to assess contour accuracy. The overall performance is measured by the average of these twi metrics, denoted as $\mathcal{J}\& \mathcal{F}$


\subsection{Implementation Detials}
We employ a 26B Sa2VA model, pretrained with InterVL2.5 and SAM2. Sa2VA is initially pretrained on image QA, video QA, image segmentation, and video segmentation datasets. For the RVOS task, Sa2VA is further trained on Ref-YouTubeVOS, MeVis, ReVOS, and a self-built Ref-SAV dataset containing 37K samples. All experiments are conducted on a single NVIDIA H100 GPU with 80GB memory.


\subsection{Main Results}

As shown in Table \ref{tab: main results}, our solution achieved a $\mathcal{J} \& \mathcal{F}$ of 64.65 on the MeVis test set, ranking 2nd place in the 7th LSVOS Challenge ROVS Track at ICCV 2025.

\setlength{\tabcolsep}{12pt} 
\begin{table}
    \centering
    \caption{The leaderboard of the MeVis test set}
    \label{tab: main results}
    \begin{tabular}{l|ccc}
    \hline
    Team & $\mathcal{J} \& \mathcal{F} $ & $\mathcal{J}$ & $\mathcal{F}$\\
    \hline
    niuqz & 67.33 & 63.82 & 70.84 \\
    \textbf{Ranhong} & 64.65 & 61.29 & 68.01 \\
    dytino & 64.14 & 61.06 & 67.22 \\
    heshuai & 62.22 & 58.99 & 65.44 \\
    DanielLi & 59.83 & 56.67 & 62.99 \\
    MVP-Lab & 57.09 & 53.32 & 60.85 \\
    \hline
    \end{tabular}
\end{table}


\subsection{Ablation Results}

\begin{figure*}[!t] 
    \centering
    \includegraphics[width=\textwidth]{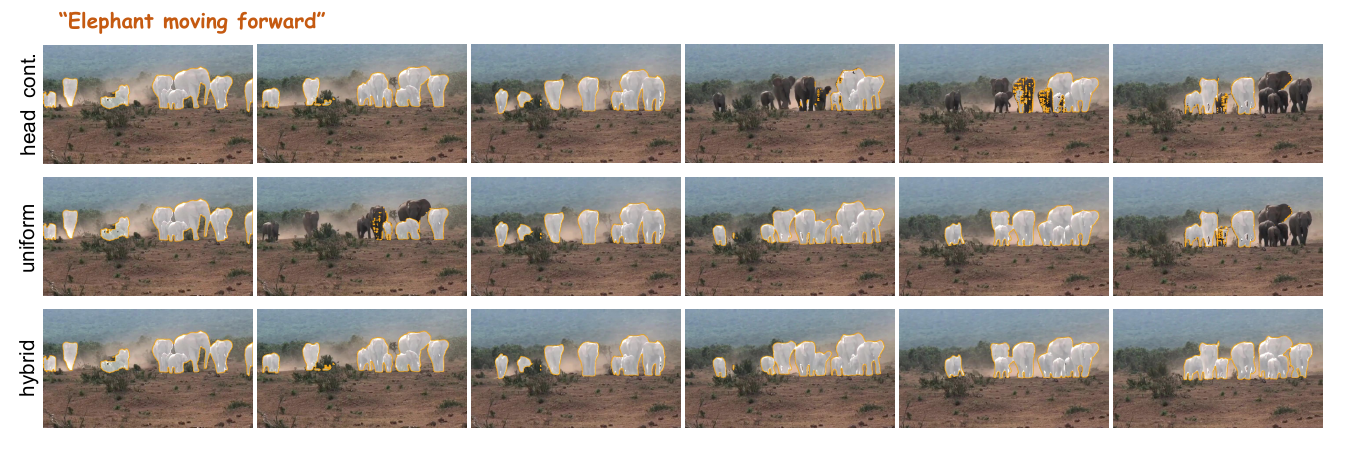}
    \caption{Ablation studies results of our method.}
    \label{fig:ablation result}
\end{figure*}

To evaluate the effectiveness of VLC and KFS, we conduct ablation studies. As shown in Table \ref{tab: ablation results}, applying KFS with the head-continue 40 frames achieves a $\mathcal{J} \& \mathcal{F} $ score of 57.67. Incorporating KFS further corrects segmentation results for mismatched video–text pairs, leading to an improved $\mathcal{J} \& \mathcal{F} $ of 61.17. Moreover, when the number of uniformly sampled frames increases from 10 to 40, the $\mathcal{J} \& \mathcal{F} $ score rises from 59.02 to 60.61, although the performance gain becomes less pronounced. 
The results obtained with different KFS strategies are illustrated in Figure \ref{fig:ablation result}. In our framework, the segmentation component of Sa2VA is implemented using SAM2, which tends to segment individual objects. Consequently, under the head-continue strategy, as the video length increases, the control effectiveness of the SEG token diminishes, leading to degraded segmentation performance for crowd-like targets.
In contrast, the uniform strategy extends the LLM-readable portion of the video to cover its entirety; however, it loses the ability to capture consecutive motion patterns, resulting in object disappearance in intermediate frames.
To address these limitations, we design a hybrid approach that combines the head-continue and uniform strategies, enabling the LLM to both access the entire video and comprehend continuous motion behaviors.

\setlength{\tabcolsep}{10pt} 
\begin{table}
    \centering
    \caption{Ablation stuides on the MeVis test set}
    \label{tab: ablation results}
    \begin{tabular}{ccc|c}
    \hline
    VLC & KFS & Number &$\mathcal{J} \& \mathcal{F} $\\
    \hline
    $\times$ & head-continue & 40 & 57.67 \\
    $\times$ & uniform & 10 & 59.02 \\
    $\times$ & uniform & 20 & 58.72 \\
    $\times$ & uniform & 30 & 59.33 \\
    $\times$ & uniform & 40 & 60.61 \\
    $\checkmark$ & head-continue & 40 & 61.17 \\
    $\checkmark$ & uniform & 40 & 64.24 \\
    $\checkmark$ & hybrid & 40 & 64.65 \\
    \hline
    \end{tabular}
\end{table}


\section{Conclusion}

In this study, we propose a method for RVOS consisting of three modules: VLC, KFS, and Sa2VA. Specifically, VLC effectively corrects segmentation errors caused by mismatches between textual and visual semantics, KFS selects the key frames that are most suitable for large models to comprehend the video content, and Sa2VA ensures high-quality video object segmentation results. Without any fine-tuning on validation or test sets, or on pseudo-labeled data, our approach 2nd place in the RVOS Track of the 7th LSVOS Challenge at ICCV 2025.
{
    \small
    \bibliographystyle{ieeenat_fullname}
    \bibliography{main}
}


\end{document}